\documentclass[11pt, a4paper, logo, copyright, nonumbering]{xiaomi}
\usepackage[authoryear, compress, round]{natbib}
\usepackage{dblfloatfix}
\usepackage{ulem}
\usepackage{caption}
\usepackage{dramatist}
\usepackage{xspace}
\usepackage{pifont} 
\usepackage{multirow}
\usepackage{tcolorbox}
\usepackage{xltabular}
\usepackage{longtable}
\usepackage{hyperref}
\usepackage{wrapfig}
\usepackage{array}
\usepackage{graphicx}

\usepackage{amsfonts}
\usepackage{amsmath}
\usepackage{amssymb}
\usepackage{lineno}
\usepackage{multirow}
\usepackage{adjustbox}

\usepackage[bottom]{footmisc}

\usepackage{CJKutf8}
\usepackage{subfigure}
\usepackage{setspace}
\usepackage{makecell}
\usepackage{graphicx}
\usepackage{subcaption}
\usepackage{multicol} 
\usepackage{siunitx}  

\newcommand{\mimobase}{MiMo-7B-Base\xspace}

\newcommand{\mimoaudioname}{MiMo-Audio\xspace}
\newcommand{\mimoaudiobase}{MiMo-Audio-7B-Base\xspace}
\newcommand{\mimoaudioinst}{MiMo-Audio-7B-Instruct\xspace}
\newcommand{\mimoaudiotokenizer}{MiMo-Audio-Tokenizer\xspace}
\newcommand{\glmvoice}{GLM-4-Voice\xspace}
\newcommand{\qwenomni}{Qwen2.5-Omni\xspace}

\newcommand{\kimiaudioinst}{Kimi-Audio-Instruct\xspace}

\newcommand{\stepaudioinst}{Step-Audio2-mini\xspace}
\newcommand{\geminiflash}{Gemini 2.5 Flash\xspace}
\newcommand{\gptaudio}{gpt-4o-audio-preview-2024-12-17\xspace}
\newcommand{\afthree}{Audio Flamingo 3\xspace}

\definecolor{xiaomiorange}{HTML}{FF6901}

\newcommand{\mytitlefont}{\fontsize{15.6}{15.6}\selectfont}

\title{\centering \mytitlefont MiMo-Audio: Audio Language Models are Few-Shot Learners}
\author{
 LLM-Core Xiaomi
}

\begin{abstract}

Existing audio language models typically rely on task-specific fine-tuning to accomplish particular audio tasks. In contrast, humans are able to generalize to new audio tasks with only a few examples or simple instructions. GPT-3 has shown that scaling next-token prediction pretraining enables strong generalization capabilities in text, and we believe this paradigm is equally applicable to the audio domain. By scaling MiMo-Audio's pretraining data to over one hundred million of hours, we observe the emergence of few-shot learning capabilities across a diverse set of audio tasks. We develop a systematic evaluation of these capabilities and find that \mimoaudiobase achieves SOTA performance on both speech intelligence and audio understanding benchmarks among open-source models. Beyond standard metrics, \mimoaudiobase generalizes to tasks absent from its training data, such as voice conversion, style transfer, and speech editing. \mimoaudiobase also demonstrates powerful speech continuation capabilities, capable of generating highly realistic talk shows, recitations, livestreaming and debates. At the post-training stage, we curate a diverse instruction-tuning corpus and introduce thinking mechanisms into both audio understanding and generation. \mimoaudioinst achieves open-source SOTA on audio understanding benchmarks (MMSU, MMAU, MMAR, MMAU-Pro), spoken dialogue benchmarks (Big Bench Audio, MultiChallenge Audio) and instruct-TTS evaluations, approaching or surpassing closed-source models. Model checkpoints and full evaluation suite are available at \textcolor{xiaomiorange} {\url{https://github.com/XiaomiMiMo/MiMo-Audio}}.
\end{abstract}

\begin{document}
\maketitle

\begin{figure}[h]
    \centering
    \begin{minipage}{\textwidth}
        \centering
        \includegraphics[width=0.9\textwidth]{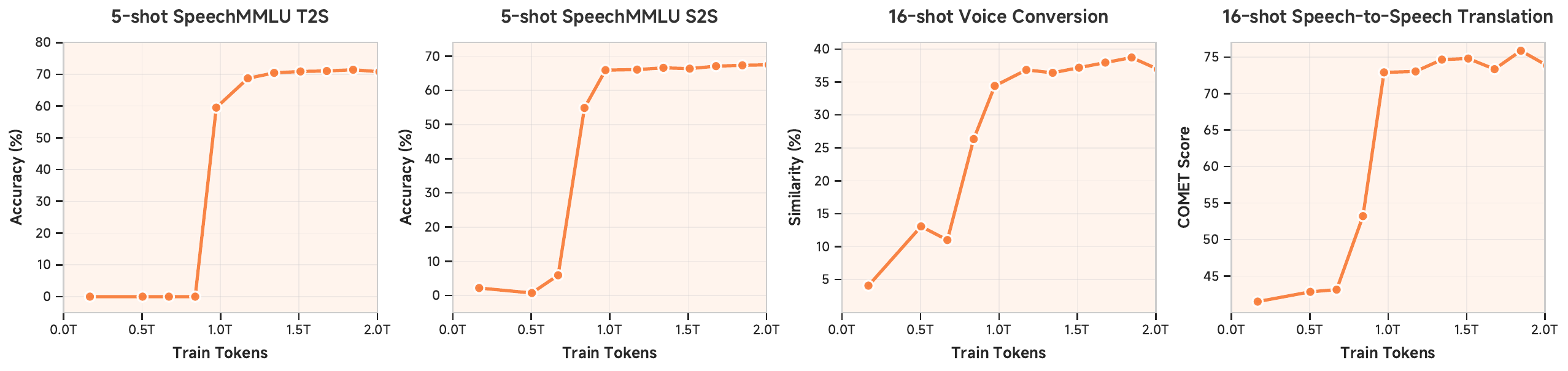}
        \label{fig:emergent}
    \end{minipage}

    \vspace{0.2cm}
    
    \begin{minipage}{\textwidth}
        \centering
        \includegraphics[width=0.85\textwidth]{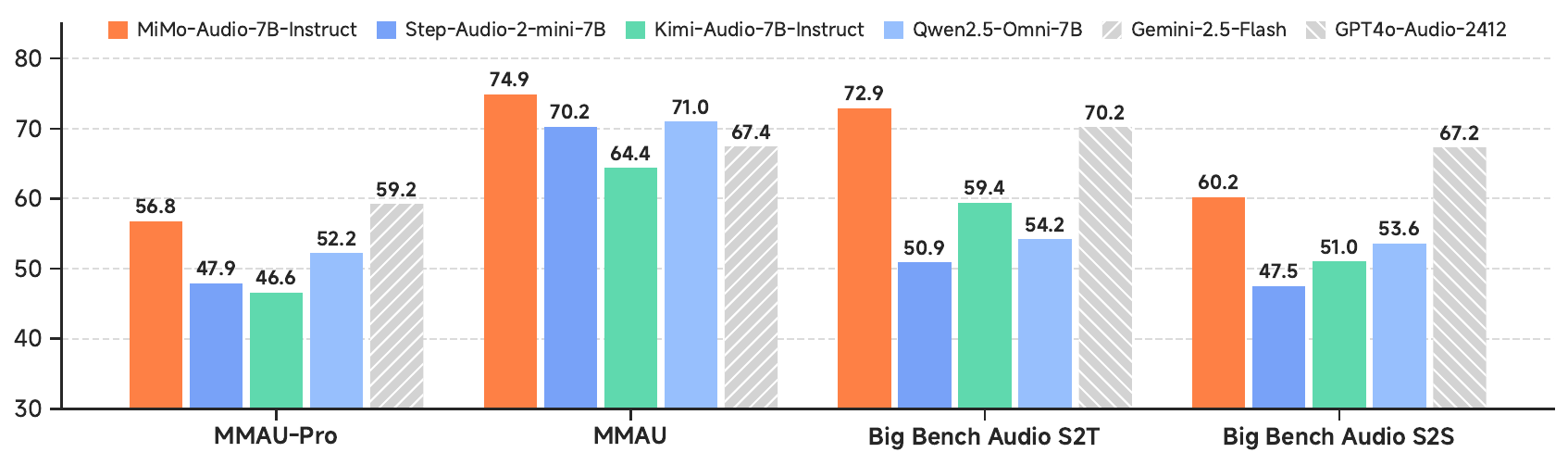}
        \caption{Emergent behavior in pretraining and performance comparison with SOTA models.}
        \label{fig:comparison}
    \end{minipage}  
\end{figure}


\newpage

\begin{spacing}{0.9}
\tableofcontents
\end{spacing}

\newpage

\section{Introduction}

Human speech interaction is characterized by its remarkable flexibility and diversity. Individuals form their understanding of speech by integrating a wide array of contextual factors—such as speakers, accents, environments, and social settings, while simultaneously modulating their own vocal expressions, like tone and prosody, in accordance with their internal states, such as mood, intent, and physical condition~\citep{SUMNER2011131, LEHET2020104328, BRADLOW2008707}. This adaptive capability is swift and dynamic, for example, humans naturally lower their voice in a quiet library or raise it during a heated debate as circumstances change.
In contrast, existing audio language models lack this inherent vocal intelligence and generalizability in comprehension and generation~\citep{zhang-etal-2023-speechgpt, kyutai2024moshi, kimi_audio, step_audio2}. To perform a range of speech tasks, including spoken dialogue, speech translation, and voice style transfer, these models still rely on being fine-tuned with task-specific datasets. 

The success of GPT-3~\citep{NEURIPS2020_1457c0d6} has proven that scaling up pre-training with next-token prediction paradigm is a viable path to achieving task generalization in the text domain. We hypothesize that this principle extends to the speech domain, where pre-training on massive-scale speech corpora using next-token prediction objective can endow a model with strong generalization abilities across a wide range of speech tasks.
While prior efforts have explored next-token prediction pretraining for speech~\citep{borsos2023audiolmlanguagemodelingapproach, zhang-etal-2023-speechgpt, kyutai2024moshi, zeng2024glm, li2025baichuan}, these models fail to achieve broad, general-purpose generalization for general speech tasks~\citep{fang2025llamaomniseamlessspeechinteraction, qwen_omni, kimi_audio, step_audio2, goel2025af3}.

We believe there are two critical aspects for next-token prediction pre-training in speech. The first is an architecture that enables the lossless flow of speech information. To fully leverage the potential of the next-token prediction paradigm, we hope all information within the speech signal to circulate through the model. This implies that we cannot use speech representations that incur a loss of paralinguistic information, which distinguishes our approach from current mainstream solutions~\citep{zeng2024glm, kimi_audio, step_audio2}.
The second aspect is scaling up. We believe that continuously scaling the volume of pre-training data will lead to sustained performance improvements and unexpected emergent abilities~\citep{wei2022emergentabilitieslargelanguage}. Therefore, we scaled our training data to over one hundred of millions of hours, which is an order of magnitude larger than the data used for the largest existing open-source speech models.
The objective of this pre-training is to equip the model with task generalization capabilities in the speech domain, meaning the model develops a broad set of atomic skills at training time, and then uses those abilities at inference time to rapidly adapt to or recognize any speech task. Our guiding principle for the pre-training method is to ensure that all information from the speech signal is preserved and flows through the model architecture.

\begin{itemize}
\item \textbf{Tokenizer}: We posit that the foremost criterion for an audio tokenizer is its reconstruction fidelity, and that its tokens should be amenable to downstream language modeling. Accordingly, we introduce \mimoaudiotokenizer. This 1.2B-parameter model employs a Transformer-based architecture comprising an encoder, a discretization layer, and a decoder, operating at a 25Hz frame rate and generating 200 tokens per second through 8 layers of residual vector quantization (RVQ). By integrating semantic and reconstruction objectives, we trained it from scratch on a 10-million-hour corpus, achieving superior performance in reconstruction quality and facilitating downstream language modeling.

\item \textbf{Architecture}: To enhance the modeling efficiency for high-token-rate (200 tokens/second) sequences and mitigate the length disparity between speech and text modalities, we propose a novel architecture combining a patch encoder, LLM, and patch decoder. The patch encoder aggregates four consecutive timesteps of RVQ tokens into a single patch, downsampling the sequence to a 6.25Hz representation for the LLM. Subsequently, the patch decoder autoregressively generates the full 25Hz RVQ token sequence.

\item \textbf{Training}: To realize a unified pre-training paradigm for both understanding and generation and to endow the model with advanced vocal intelligence, we devise a two-stage training strategy, leveraging \mimobase~\citep{coreteam2025mimounlockingreasoningpotential} for initialization. Stage 1 is dedicated to speech understanding, while stage 2 integrates both understanding and generation in a unified framework. Each stage features tailored training tasks. Notably, we observed the spontaneous emergence of in-context learning abilities for speech during this process.

\item \textbf{Data}: We have scaled our pre-training corpus to an unprecedented over 100 million hours of speech data, representing an order-of-magnitude increase over any existing open-source speech model. This was supported by a purpose-built, end-to-end data pipeline for pre-processing, annotation, and curation.

\item \textbf{Evaluation}: We have developed a comprehensive benchmark to rigorously assess the model's in-context learning capabilities in the speech domain. The benchmark is designed to evaluate multiple facets, including modality-invariant general knowledge, auditory comprehension and reasoning, and a diverse suite of speech-to-speech generation tasks.


\end{itemize}


After large-scale pre-training, \mimoaudiobase demonstrates strong few-shot learning capabilities~\citep{brown2020languagemodelsfewshotlearners}. It exhibits very high "Speech Intelligence" and strong modality alignment when evaluated on our constructed SpeechMMLU, which originates from MMLU~\citep{hendrycks2021measuringmassivemultitasklanguage} and is built by synthesizing its tasks into speech. \mimoaudiobase achieves superior performance under speech input and output, with results closely approaching text-based MMLU, and incurs only a minor degradation in text performance. It also shows excellent generalization to unseen tasks: with just a few demonstrations in the context, it can perform tasks such as voice conversion, style transfer, speech rate control, denoising, and speech translation. Furthermore, \mimoaudiobase displays powerful speech continuation abilities, generating highly realistic and semantically coherent monologues or multi-speaker dialogues in formats like talk shows, speeches, debates, podcasts, and game commentaries.

We believe the core objective of post-training is to align the model's pre-trained generalization capabilities with instruction-following abilities. To this end, we construct a highly diverse instruction-tuning corpus for audio understanding and generation by aggregating high-quality open-source and in-house data spanning multiple domains. To further enhance the model's cross-modal reasoning abilities, we also created high-quality "thinking"~\cite[chain-of-thought;][]{wei2022chain} data for both audio understanding and generation tasks. To obtain human-like and style-controllable speech dialogue data, we trained MiMo-TTS-7B on over 7 million hours of data to convert text-based conversations into speech. \mimoaudioinst demonstrates superior audio understanding and reasoning abilities after post-training. It achieves SOTA results among open-source models on audio understanding/reasoning benchmarks such as 
MMSU~\citep{wang2025mmsumassivemultitaskspoken},
MMAU~\citep{sakshi2025mmau}, MMAR~\citep{ma2025mmarchallengingbenchmarkdeep}, and MMAU-Pro~\citep{kumar2025mmauprochallengingcomprehensivebenchmark}, approaching or surpassing the performance of closed-source models. \mimoaudioinst also shows exceptional speech intelligence and instruction-following capabilities, significantly outperforming other open-source models on spoken dialogue benchmarks like Big Bench Audio and MultiChallenge Audio~\citep{sirdeshmukh2025multichallengerealisticmultiturnconversation}. In instruction-following TTS tasks, its performance is comparable to that of GPT-4o-mini-tts.

Our key contributions are:
\begin{itemize}
\item  We present the first empirical evidence that scaling lossless, compression-based speech pre-training to an unprecedented 100 million hours unlocks emergent task generalization, exemplified by powerful few-shot learning abilities. We argue this represents a  "GPT-3 moment" for the speech domain.

\item  We propose the first comprehensive and replicable blueprint for generative speech pre-training, which includes a novel tokenizer, a scalable architecture, a phased training strategy, and a holistic evaluation suite. 

\item  We pioneer the integration of thinking into the modeling process for both speech understanding and generation, bridging the gap between perception and complex cognitive tasks.

\end{itemize}

\section{Model Architecture}

\subsection{\mimoaudiotokenizer}
A main challenge in existing audio tokenization methods lies in effectively balancing the inherent trade-off between semantic and acoustic information in audio signals. Semantic tokens, typically derived from self-supervised learning models~\citep{hsu2021hubert,chung2021w2v,zhang2023google} or ASR models~\citep{zeng2024glm,li2025baichuan}, exhibit a strong correlation with linguistic content, facilitating alignment with the text modality. However, their primary drawback is the loss of fine-grained acoustic information, which constrains the quality of raw waveform reconstruction. In contrast, acoustic tokens generated by neural audio codecs~\citep{zeghidour2021soundstream, defossez2022high} enable high-fidelity audio reconstruction but struggle to establish effective alignment with the text semantic space. 

To jointly capture both semantic and acoustic information, prior works such as SpeechTokenizer~\citep{zhang2023speechtokenizer} and Mimi~\citep{kyutai2024moshi} have attempted to incorporate semantic distillation strategies into neural audio codecs to obtain unified audio tokens. Nevertheless, constrained by the limited scale of their encoders, these methods struggle to fully mitigate the conflict between semantic and acoustic information and their semantic expressiveness remains inferior to semantic tokens. Other approaches, like X-Codec~\citep{ye2025codec} and XY-Tokenizer~\citep{gong2025xy}, employ a dual-stream architecture with separate semantic and acoustic encoders to alleviate these issues. However, these methods still rely on pre-trained semantic models, and their dual-encoder architecture results in semantic and acoustic information originating from separate representation spaces.

To address these limitations, we propose \mimoaudiotokenizer, a unified tokenizer trained from scratch that is capable of both capturing semantic information and enabling high-fidelity audio reconstruction. By scaling up the model's parameters and training data, \mimoaudiotokenizer further alleviates the semantic-acoustic representation conflict, thereby enhancing both cross-modal alignment and speech reconstruction quality.

\begin{figure}[t]
    \centering
    \includegraphics[width=1.0\textwidth]{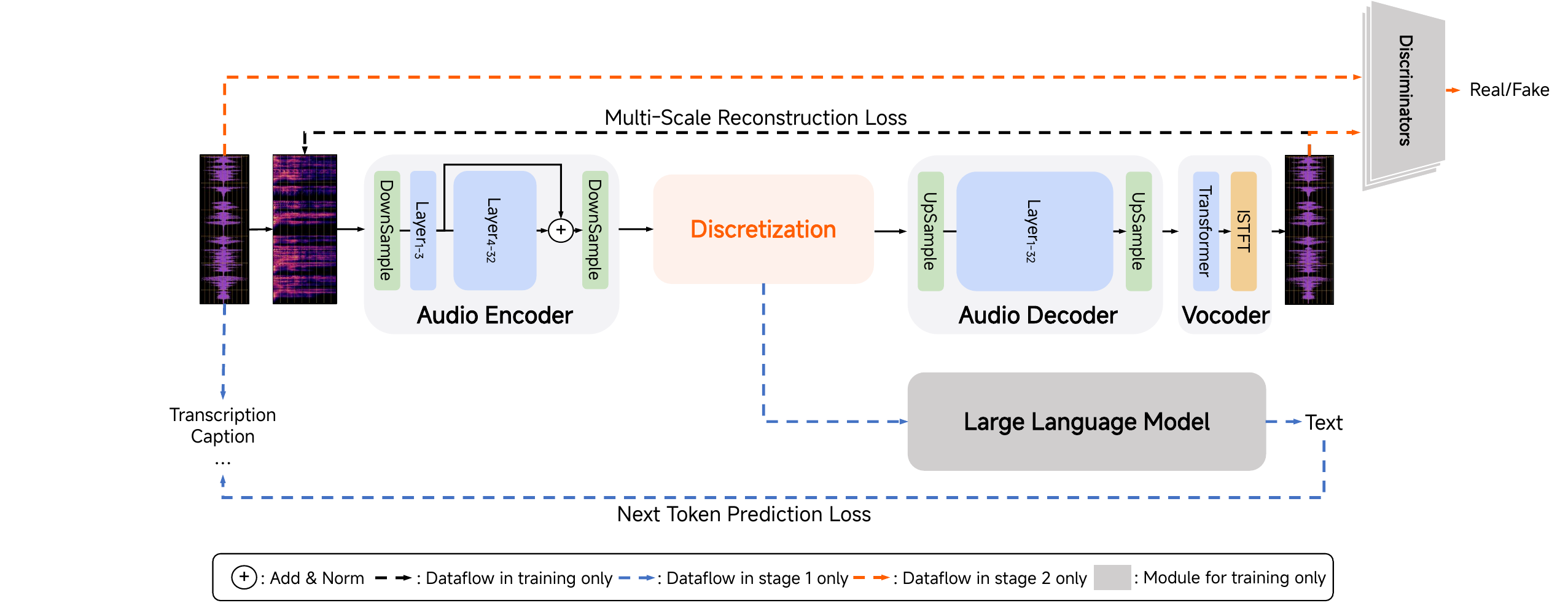}
    \caption{
    Illustration of~\mimoaudiotokenizer framework.
    }
    \label{fig:mimo_audio_tokenizer}
\end{figure}
\subsubsection{Architecture}
As illustrated in Figure \ref{fig:mimo_audio_tokenizer}, the architecture of \mimoaudiotokenizer comprises four main components: an audio encoder, a discretization module, an audio decoder and a vocoder. The audio encoder is composed of a central Transformer encoder with bidirectional attention, bracketed by $2\times$ downsampling layers at the input and output. The central encoder consists of 32 layers with 20 attention heads, employing Rotary Position Embeddings~\cite[RoPE;][]{su2024roformer} and GELU activations~\citep{hendrycks2016gaussian}. We set the model dimension to 1280 and the FFN inner dimension to 5120. To mitigate the conflict between semantic and acoustic information, we add the layer-3 hidden states to the final-layer output via element-wise summation. The discretization module has a 20-layer Residual Vector Quantizer~\cite[RVQ;][]{oord2018neuraldiscreterepresentationlearning,zeghidour2021soundstream}, where the first two layers have a codebook size of 1024, and the remaining layers use a size of 128.  The audio decoder adopts a mirror structure to the encoder but employs causal self-attention to support streaming generation. The vocoder follows the Vocos design~\citep{siuzdak2024vocosclosinggaptimedomain} but replaces the ConvNeXt~\citep{liu2022convnet2020s} backbone with a Transformer, enabling sequence packing for more efficient training. The Transformer has 16 layers, 16 heads, a model dimension of 256, an FFN dimension of 1024. It incorporates RoPE and sliding window attention with window sizes of [40, 10], which provides the Vocoder with receptive fields of [6.4s, 1.6s].

Given a single-channel audio waveform $X$ sampled at 24 kHz, we first convert it into a mel-spectrogram with a frame rate of 100 Hz. This spectrogram is then fed into the audio encoder, which transforms it into a sequence of continuous representations of length $M$ at 25 frame rate. The RVQ within the discretization module subsequently quantizes these continuous representations into a 2D matrix of discrete indices $A \in \mathbb{N}^{M \times R}$, where $R$ is the number of RVQ layers. These indices are then used to reconstruct the quantized representation $\mathbf{Q}$ by looking up and summing the corresponding embeddings from the codebooks. Finally, the audio decoder and the vocoder reconstruct the audio waveform $\hat{X}$ from $\mathbf{Q}$.

\subsubsection{Training}

Inspired by \citet{wu2023audiodec}, we employ a two-stage training paradigm to enhance training efficiency as depicted in Figure \ref{fig:mimo_audio_tokenizer}. In stage 1, the model undergoes multi-task learning on a large-scale dataset. Specifically, we scale up the training data to over 11 million hours. This extensive training enables the model to jointly encode both semantic and acoustic information. In stage 2, the parameters of the audio encoder and discretization module are frozen. Discriminators are introduced to train the audio decoder and vocoder, focusing on improving the reconstruction of fine-grained details in the original audio waveform and eliminating vocoding artifacts.

\paragraph{{Unified Representation Learning}}
In stage 1, we combine the audio reconstruction task and the audio-to-text (A2T) task to align the representation spaces of audio and text while ensuring the preservation of acoustic information. To provide supervision for the A2T objective, we introduce an LLM that is jointly trained with \mimoaudiotokenizer. All parameters of \mimoaudiotokenizer and LLM are trained from scratch. We formulate the A2T objective as a next-token prediction loss applied to the LLM's text output, defined as:
\begin{align}
    \mathcal{L}_{\text{A2T}}=-\sum_{i=1}^N\log{p(t_i|\tilde{\mathbf{Q}}, t_1, \dots,t_{i-1})},
\end{align}
where $T=[t_1,\dots,t_N]$ is the target text sequence, $\tilde{\mathbf{Q}}$ is the quantized audio representation, and $N$ is the total length of the text sequence.

For the audio reconstruction task, we adopt a multi-scale mel-spectrogram reconstruction loss, defined as the $L1$ distance:
\begin{align}
      \mathcal{L}_{\text{recon}}=\sum_{i\in e}\lVert \mathcal{S}_i(X)-\mathcal{S}_i(\hat{X})\lVert_1,
\end{align}
where $\mathcal{S}_i$ denotes the mel-spectrogram at scale $i$ with $2^i$ bins, computed using a normalized Short-Time Fourier Transform (STFT) with a window size of $15 \cdot 2^{i-1}$ and a hop length of $15 \cdot 2^{i-2}$. The set of scales is defined as $e=\{5,6,7\}$. Finally, including the commitment loss $\mathcal{L}_{\text{commit}}$ from the discretization module, the total loss for stage 1 is a weighted sum:
\begin{align} \mathcal{L}_{\text{stage1}}=\lambda_{\text{A2T}}\mathcal{L}_{\text{A2T}}+\lambda_{\text{recon}}\mathcal{L}_{\text{recon}}+\lambda_{\text{commit}}\mathcal{L}_{\text{commit}},
\end{align}
where $\lambda_{\mathrm{A2T}}{=}10.0,\;
\lambda_{\mathrm{recon}}{=}1.0,\;
\lambda_{\mathrm{commit}}{=}1.0$.

\paragraph{Adversarial Fine-tuning}
In stage 2, we introduce additional discriminators for adversarial training to improve waveform reconstruction quality. During this stage, all parameters involved in the audio tokenization process are frozen to preserve the semantic structure of the audio token space. We adopt a multitask GAN training recipe that jointly optimizes (i) a mel-spectrogram reconstruction loss from stage~1, (ii) an adversarial loss, and (iii) a discriminator feature-matching loss. To provide supervision in both the time and frequency domains, we employ a Multi-Period Discriminator \cite[MPD;][]{kong2020hifigangenerativeadversarialnetworks} together with a Multi-Scale STFT discriminator \cite[MS-STFT;][]{défossez2022highfidelityneuralaudio}. We adopt the Hinge-GAN \citep{lim2017geometricgan,miyato2018spectralnormalizationgenerativeadversarial} training framework, applying spectral normalization to all discriminator layers and disabling weight decay during discriminator training. Let $\mathcal{D}=\{D_k\}_{k=1}^{K}$ denote the full set of sub-discriminators across MPD and MS-STFT. Given a real waveform $X$ and a generated waveform $\hat{X}$, the discriminator objective can be formulated as
\begin{equation}
\mathcal{L}_{D}
=\frac{1}{K}\sum_{k=1}^{K}
\Big[
\mathbb{E}_{X}\big[\max(0,\,1 - D_k(X))\big]
+
\mathbb{E}_{\hat{X}}\big[\max(0,\,1 + D_k(\hat{X}))\big]
\Big],
\end{equation}
and the generator adversarial objective is
\begin{equation}
\tilde{\mathcal{L}}_{\mathrm{adv}}
= -\,\frac{1}{K}\sum_{k=1}^{K}\mathbb{E}_{\hat{X}}\big[D_k(\hat{X})\big],
\end{equation}
where the normalization by $\tfrac{1}{K}$ prevents the number of sub-discriminators from dominating the optimization.
For feature matching, we minimize the $\ell_1$ distance between intermediate discriminator activations:
\begin{equation}
\mathcal{L}_{\mathrm{fm}}
= \frac{1}{K}\sum_{k=1}^{K}\frac{1}{L_k}\sum_{\ell=1}^{L_k}
\big\| f_{k,\ell}(X)-f_{k,\ell}(\hat{X}) \big\|_{1},
\end{equation}
where $f_{k,\ell}(\cdot)$ returns the $\ell$-th–layer features of $D_k$, and $L_k$ denotes the number of intermediate layers included. When forming the composite objective, we assign fixed weights to the individual losses to keep their gradient magnitudes on comparable scales. The generator is trained with
\begin{equation}
\mathcal{L}_{G}
= \lambda_{\mathrm{recon}}\,\mathcal{L}_{\mathrm{recon}}
+ \lambda_{\mathrm{adv}}\,\tilde{\mathcal{L}}_{\mathrm{adv}}
+ \lambda_{\mathrm{fm}}\,\mathcal{L}_{\mathrm{fm}},
\quad
\end{equation}
where $\lambda_{\mathrm{recon}}{=}1.0,\;
\lambda_{\mathrm{adv}}{=}1.0,\;
\lambda_{\mathrm{fm}}{=}2.0$.

\subsubsection{Evaluation}

\paragraph{Settings} 
We assess the preservation of acoustic information in audio tokenization with multiple metrics. These include: Speaker Similarity (SIM), calculated as the cosine similarity of embeddings from a pre-trained speaker verification model\footnote{\url{https://github.com/microsoft/UniSpeech/tree/main/downstreams/speaker_verification}}; Short-Time Objective Intelligibility~\cite[STOI;][]{taal2010short}; and Perceptual Evaluation of Speech Quality~\cite[PESQ;][]{rix2001perceptual}. All evaluations are conducted on the ground-truth recordings of Seed-TTS-Eval~\citep{anastassiou2024seed}. The compared baselines include GLM-4-Voice-Tokenizer~\citep{zeng2024glm}, Baichuan-Audio-Tokenizer~\citep{li2025baichuan}, XY-Tokenizer~\citep{gong2025xy}, Mimi~\citep{kyutai2024moshi}, XCodec~\citep{ye2025llasascalingtraintimeinferencetime}, and BigCodec~\citep{xin2024bigcodec}.
Considering our downstream \mimoaudioname is trained exclusively on audio tokens produced by the first eight codebooks of \mimoaudiotokenizer, we evaluate and compare waveform reconstruction quality decoded using only those codebooks. This protocol faithfully reflects the fidelity of the audio accessible to the downstream language model. We evaluate Mimi under the same protocol for consistency.

\paragraph{Results}
As shown in Table~\ref{tab:tokenizer}, \mimoaudiotokenizer delivers strong reconstruction quality on Seed-TTS-Eval. Across both ZH and EN splits, it achieves the highest scores on PESQ-NB/WB, SIM, and STOI, substantially outperforming all baselines at a comparable bitrate. Crucially, these gains are measured exactly on the codebooks used for downstream modeling, indicating that \mimoaudioname preserves the full fidelity of speech information, which in turn yields strong generalization across diverse speech tasks.

\begin{table}[t]
\centering
\small
\setlength{\tabcolsep}{4pt}
\resizebox{\linewidth}{!}{
\begin{tabular}{
  l c |
  S[table-format=1.2] S[table-format=1.2] S[table-format=1.2] S[table-format=1.2] |
  S[table-format=1.2] S[table-format=1.2] S[table-format=1.2] S[table-format=1.2]
}
\toprule
&&\multicolumn{4}{c|}{\textbf{SEED-ZH}}&\multicolumn{4}{c}{\textbf{SEED-EN}}\\

\textbf{System} & \textbf{kBPS} &
\textbf{PESQ-NB} & \textbf{PESQ-WB} & \textbf{SIM} & \textbf{STOI} & 
\textbf{PESQ-NB} & \textbf{PESQ-WB} & \textbf{SIM} & \textbf{STOI}  \\
\midrule

\mimoaudiotokenizer    & 1.55    & \textbf{3.30} & \textbf{2.71} & \textbf{0.89} & \textbf{0.93}  & \textbf{3.02} & \textbf{2.43} & \textbf{0.85} & \textbf{0.92} \\
 GLM-4-Voice-Tokenizer & 0.175 &1.11 & 1.06 & 0.33&0.61&1.11&1.05&0.12&0.60\\
Baichuan-Audio-Tokenizer& 1.0     &  2.37    &  1.84   &   0.78   &  0.86      &   2.11   &   1.62   & 0.69     &   0.85       \\
XY-Tokenizer        & 1.0     &  2.88   &   2.24   & 0.87     &0.90         &  2.69    &  2.14    & 0.82     &   0.90     \\
Mimi               & 1.1      & 2.57 & 2.05 & 0.73 & 0.88   & 2.60 & 2.07 & 0.74 & 0.89 \\
XCodec2.0        & 0.8      &  2.69    &   2.10   &   0.81   &   0.89      & 2.57     & 2.01    &  0.78    &    0.89     \\  
BigCodec&1.04&2.88&2.26&.80&.91&2.80&2.22&0.80&0.91\\
\bottomrule
\end{tabular}
}
\caption{Evaluation of audio tokenizers on Seed-TTS-Eval dataset. ZH/EN split results are reported in the same row for each system. kBPS denotes the effective bitrate (kilobits per second) of the tokenized audio stream.}
\label{tab:tokenizer}
\end{table}

\subsection{\mimoaudioname}

\begin{figure}[t]
    \centering
    \includegraphics[width=1.0\textwidth]{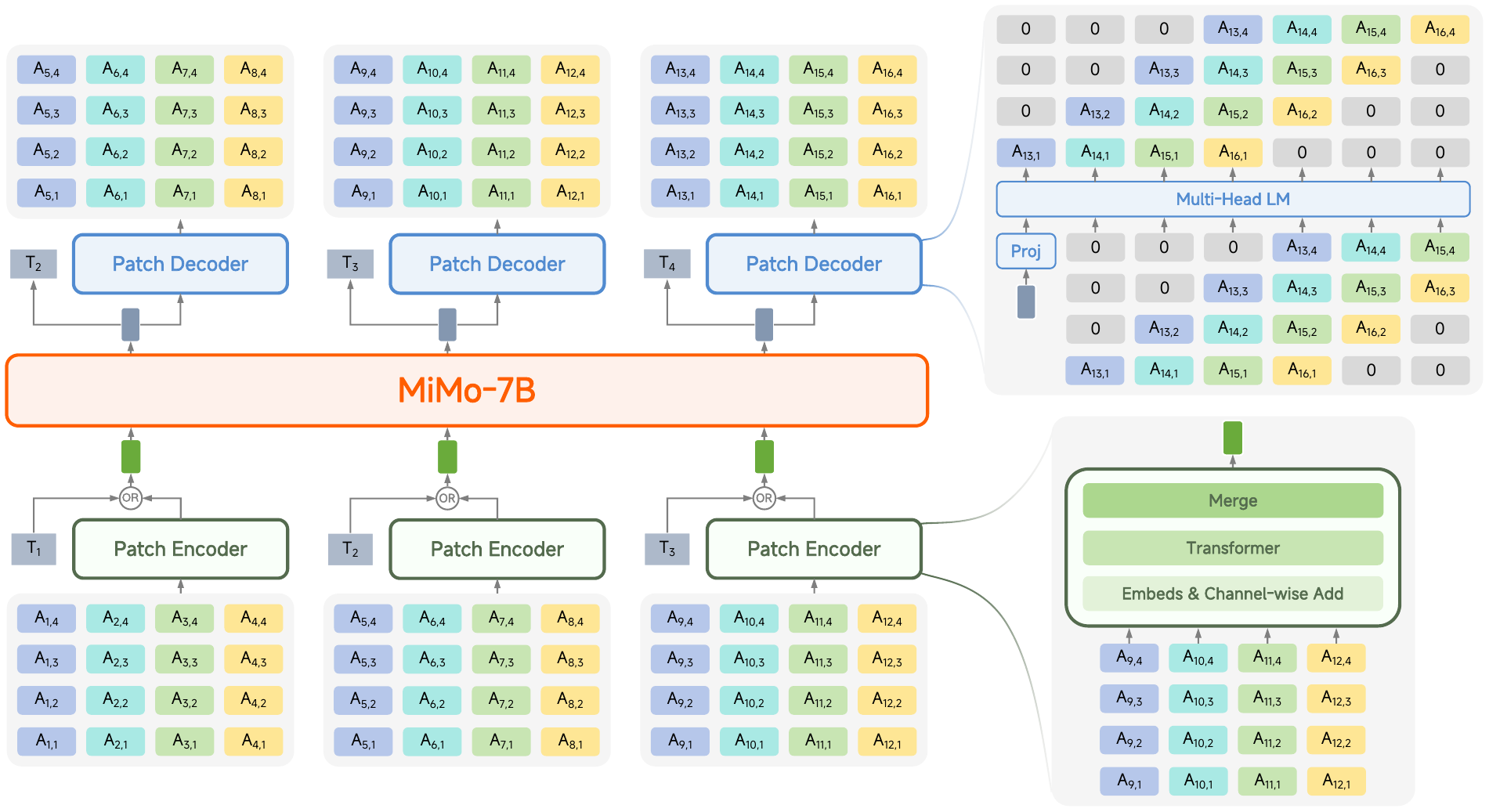}
    \caption{
    Model architecture of~\mimoaudioname.
    }
    \label{fig:mimo_audio}
\end{figure}

\mimoaudioname is a unified generative audio-language model that jointly models sequences of text and audio tokens, as illustrated in Figure~\ref{fig:mimo_audio}. The model accepts both text and audio tokens as input and autoregressively predicts either text or audio tokens, thereby supporting a comprehensive range of tasks involving arbitrary combinations of text and audio modalities. 

Formally, let $T=[t_1,\ldots,t_N]$ denote the text sequence and the audio token sequence be defined as:
\begin{align} 
A=[A_1, \ldots, A_M],\qquad
A_i\triangleq(a_{i,1},\ldots,a_{i,R'}),
\end{align}
where $N$ denotes the text sequence length, $M$ the audio sequence length, and $R'=8$ the number of RVQ codebooks used for LLM training. Since audio sequences have relatively low information density, individual audio frames convey much less information than text tokens. To mitigate this mismatch in granularity across modalities and facilitate cross-modal knowledge transfer, we partition the audio sequence into contiguous groups of $G$ frames, forming \textbf{audio patches}:
\begin{align} 
P=[P_1,\ldots,P_{M/G}],\qquad
P_i=[A_{(i-1)G+1}, \ldots, A_{iG}].
\end{align}
The input to \mimoaudioname is the interleaved sequence of text tokens and audio patches. Let $S=[s_1, \ldots, s_L]$ denote the interleaved sequence, where each element $s_i$ is either a text token or an audio patch. The model is trained autoregressively:
\begin{align} 
p(S) = \prod_{i=1}^{L} p(s_i | s_1, \dots, s_{i-1}),
\end{align}
where $p(s_i | s_1, \dots, s_{i-1})$ represents next-token prediction when $s_i$ is a text token or next-patch prediction when $s_i$ is an audio patch. This unified modeling approach enables seamless handling of arbitrary text-audio interleaved sequences. \mimoaudioname comprises three primary components: a patch encoder, an LLM backbone, and a patch decoder, which we describe in detail below.

\subsubsection{Patch Encoder}

The patch encoder transforms audio tokens within each patch into a single hidden vector. We maintain $R'$ distinct embedding tables $\{E_r\}_{r=1}^{R'}$ that map audio tokens to their corresponding embedding vectors. For each audio token $a_{i,r}$, we obtain its embedding as $\mathbf{e}_{i,r}=E_r(a_{i,r})$. The embeddings across all RVQ codebooks for frame $i$ are aggregated to form a unified representation:
\begin{align}
\mathbf{e}_i=\sum_{r=1}^{R'}\mathbf{e}_{i,r}.
\end{align}
The resulting sequence within each patch is processed by a Transformer encoder with $L_{\mathrm{enc}}=6$ layers. Each layer has a hidden dimension of 1024, 64 attention heads, and an FFN dimension of 4096. The encoder employs bidirectional self-attention, which enables the model to capture local contextual information across frames. The outputs from all frames within the patch are subsequently concatenated and projected through a linear transformation layer to match the input dimensionality of the LLM.

\subsubsection{Large Language Model}

We employ \mimobase~\citep{coreteam2025mimounlockingreasoningpotential} as the LLM backbone. The model accepts inputs at each position as either text token embeddings or audio patch representations produced by the patch encoder. The resulting hidden states can be processed through an output projection layer for text token prediction or fed to the patch decoder for audio patch generation, as described in the subsequent section.

\subsubsection{Patch Decoder}

The patch decoder autoregressively generates audio tokens within each patch during audio generation. It comprises $L_{\mathrm{dec}}=16$ Transformer layers, each with a hidden dimension of 1024, 64 attention heads, and an FFN dimension of 4096. The decoder employs causal masking in the self-attention mechanism. The patch decoder employs the same $R'$ embedding tables as the patch encoder, one for each RVQ codebook. To facilitate RVQ token generation, the Transformer is equipped with $R'$ independent output heads, each dedicated to predicting tokens for a specific RVQ codebook.

Formally, given a hidden state $\mathbf{h}$ from the LLM, let $P=[A_{1}, \ldots, A_{G}]$ denote the audio patch to be generated. The naive approach involves autoregressive generation of audio frames within each patch along the temporal dimension:
\begin{align}
p(P|\mathbf{h}) = \prod_{i=1}^{G} p(A_{i}|\mathbf{h}, A_1, \dots, A_{i-1}),
\end{align}
where the probability for each frame $A_i$ decomposes across the $R'$ codebooks:
\begin{align}
p(A_{i}|\mathbf{h}, A_1, \dots, A_{i-1}) = \prod_{r=1}^{R'} p(a_{i,r}|\mathbf{h}, A_1, \dots, A_{i-1}).
\end{align}
However, due to dependencies between tokens across different RVQ layers, predicting all RVQ tokens simultaneously at each time step is challenging and often leads to poor audio generation quality. To mitigate this limitation, we introduce a delay mechanism for audio token generation, inspired by~\citet{musicgen}. Specifically, we introduce layer-specific delays $D=[d_1, \ldots, d_{R'}]$, where $d_r$ represents the delay (in time steps) for generating tokens at RVQ layer $r$. The delayed audio patch is formalized as:
\begin{align}
P'=[A'_{1}, \ldots, A'_{G+\max(D)}],
\end{align}
where 
\begin{align}
a'_{i,r} = \begin{cases}
a_{i-d_r,r} & \text{if } 1 \leq i - d_r \leq G \\
0 & \text{otherwise}
\end{cases}
\end{align}
for $i \in [1, G+\max(D)]$ and $r \in [1, R']$. Here, 0 denotes an empty token that is disregarded during both encoding and decoding processes. The patch decoder models these delayed audio patches autoregressively following the aforementioned formulation and maintains the delay pattern during the decoding phase. We list the detailed model configuration in Table~\ref{tab:model_config}.

\begin{table}[tbp]
\centering
\begin{tabular}{l|>{\centering\arraybackslash}m{3.cm}|
                  >{\centering\arraybackslash}m{3.cm}|
                  >{\centering\arraybackslash}m{3.cm}}
\toprule
\textbf{Hyper-parameter} & \textbf{Patch Encoder} & \textbf{LLM} & \textbf{Patch Decoder} \\
\midrule
\multicolumn{4}{c}{\textbf{Model Architecture}} \\
\midrule
model dimension   & 1024 & 4096 & 1024 \\
FFN dimension     & 4096 & 11008 & 4096 \\
attention heads   & 64   & 32   & 64 \\
number of layers  & 6    & 36   & 16 \\
context length    & 4    & 8192 & 11 \\
\midrule
\multicolumn{4}{c}{\textbf{Input/Output Space}} \\
\midrule
text vocab size   & \multicolumn{3}{c}{151680} \\
audio channels    & \multicolumn{3}{c}{8} \\
audio vocab sizes & \multicolumn{3}{c}{1024-1024-128-128-128-128-128-128} \\
audio frame rate  & \multicolumn{3}{c}{6.25 Hz} \\
\bottomrule
\end{tabular}
\caption{Model architecture and Input/Output space configuration.}
\label{tab:model_config}
\end{table}

\section{Pre-Training}

\subsection{Data}
Our pre-training corpus consists of unimodal data (text-only and speech-only) and multimodal data (speech–text).
The construction procedure for the text-only corpus is described in MiMo~\citep{coreteam2025mimounlockingreasoningpotential}.
For the speech modality, the objective is to provide the model with large-scale, high-quality, and diverse audio data.
To this end, we developed a comprehensive data pipeline that integrates data collection, automated processing, multi-dimensional annotation, and quality control.

\subsubsection{Data Preprocessing}
Our pre-training data contains hundreds of millions of hours of In-the-wild audio data, and we ensure the data's diversity in terms of source and content.

\begin{itemize}
    \item \textbf{Source Diversity}: The data covers a variety of sources, such as public podcasts, audiobooks, news broadcasts, interviews, and conference recordings. This multi-source, heterogeneous data combination ensures the model will not be biased towards specific recording environments or speaking styles.
    \item \textbf{Content Diversity}: The data covers topic areas such as daily communication, entertainment media, business and entrepreneurship, arts and culture, and scientific research. This enables the model to learn about rich knowledge domains.
\end{itemize}

To transform large-scale raw audio into high-quality training data, we designed and implemented an efficient and scalable automated pipeline, inspired by previous work~\citep{yu2023autoprepautomaticpreprocessingframework, he2024emiliaextensivemultilingualdiverse, kang2024libriheavy50000hoursasr, song2024touchttsembarrassinglysimpletts}.
The pipeline incorporates modules such as audio normalization, speaker diarization, voice activity detection (VAD), automatic speech recognition (ASR), and audio quality assessment.

\subsubsection{Data Labeling}
To accurately evaluate and filter the pre-training data, we built an automated annotation system covering semantic and non-semantic dimensions to generate rich, structured attribute labels for each piece of data.

\begin{itemize}
    \item \textbf{Semantic Dimension}: Based on the transcription results from modules like ASR, we built a text quality assessment model. This model can score the semantic value of the content from multiple dimensions such as conversational quality, knowledge density, and logical reasoning.

    \item \textbf{Non-semantic Dimension}: To obtain non-semantic level information, we trained an audio captioning model. This model can directly generate rich natural language descriptions for the audio (such as non-semantic information like timbral characteristics, emotional style, and background environment).
\end{itemize}

This dual-dimension annotation method not only measures data quality but also endows the corpus with more fine-grained attribute information, thereby supporting more efficient and targeted filtering and training.

\subsubsection{Data Curation}
On the basis of multi-dimensional data annotation, we conducted rigorous filtering and sampling of the data.
\begin{itemize}
    \item \textbf{Low-Quality Data Filtering}: According to preset quality thresholds, we removed segments containing excessive noise, low-quality, and unsafe content, ensuring the reliability of the final corpus.

    \item \textbf{High-Quality Data Sampling}: We integrated scoring metrics from semantic and non-semantic dimensions and designed a sampling strategy to ensure the model can learn efficiently from the high-quality corpus.
\end{itemize}

\subsection{Training}

\begin{table}[tbp]
\centering
\begin{tabular}{l|c|c|c}
\toprule
\multirow{2}{*}{\textbf{Hyper-parameter}} & \multicolumn{2}{c|}{\textbf{Pre-training}} & \multirow{2}{*}{\textbf{Post-training}} \\
\cline{2-3}
 & \textbf{Understanding} & \textbf{Understanding-Generation} &  \\
\midrule
LR (Patch Encoder) & 2e-4 & 2e-4 & 5e-5 \\
LR (LLM)           & 3e-5 & 3e-5 & 1e-5 \\
LR (Patch Decoder) & -    & 2e-4 & 5e-5 \\
LR scheduler                  & constant & cosine & cosine \\
batch size                    & 16.8M tokens & 16.8M tokens & 2.1M tokens \\
warmup ratio                  & 0.01 & 0.01 & 0.01 \\
loss weights                  & 1-0-0-0-0-0-0-0-0 & 100-12-8-6-4-2-2-1-1 & 100-12-8-6-4-2-2-1-1 \\
delay patterns  & - &  0-1-2-3-4-5-6-7 & 0-1-2-3-4-5-6-7 \\
\bottomrule
\end{tabular}
\caption{Training configuration across different stages. LR stands for learning rate.}
\label{table:training_config}
\end{table}

Our training starts from the \mimobase model. To maximally preserve its text capabilities while simultaneously equipping the model with speech understanding and generation abilities, MiMo-Audio employs a progressive, two-stage pre-training method.

\subsubsection{Understanding Training}
In the first stage, we train the model's patch encoder and LLM components. This stage aims to enable the model to master speech understanding capabilities. We constructed a dataset of 2.6T tokens in total, consisting of 1.2T text tokens and 1.4T speech-related tokens~(calculated at a 6.25Hz speech frame rate). The data includes four task formats: speech-text interleaved data, ASR data, general audio captioning data, and text-only pre-training data. During this stage, we only compute the loss on the text tokens. As detailed in the Table~\ref{table:training_config}, the learning rate for the patch encoder is 2e-4, while the LLM's learning rate is 3e-5, with a constant learning rate scheduler. Each batch contains 16.8M tokens, and the training context length is 8192.

\subsubsection{Understanding-Generation Joint Training}
In the second stage, we train all parameters of the model, including the patch encoder, LLM, and patch decoder. This stage is designed to provide the model with an integrated capability for both speech understanding and generation. The training dataset has 5T tokens, comprising 2.6T text tokens and 2.4T audio tokens~(calculated at a 6.25Hz speech frame rate). This includes seven task formats: speech continuation, speech-text interleaved data, ASR, TTS, general audio captioning, instruction-following TTS, and text pre-training data. For tasks that require speech generation, such as speech continuation or generating the speech segments within speech-text interleaved and TTS data, we employ a text-guided interleaving generation strategy to improve speech generation quality. Specifically, the model interleaves text tokens and speech patches in a fixed 5:5 ratio. Once text generation is complete, the model generates the remaining speech tokens until completion. In this stage, we compute the loss on both text and audio tokens. The loss weight for text tokens is 100, while the weights for the respective RVQ tokens are 12, 8, 6, 4, 2, 2, 1, and 1. As shown in the Table~\ref{table:training_config}, the learning rate for the patch encoder and decoder is 2e-4, the LLM's learning rate is 3e-5, and the learning rate scheduler follows a cosine decay. The batch size and context length remain consistent with Stage 1.

\subsection{Evaluation}

We evaluate \mimoaudiobase using two types of evaluation: few-shot in-context learning evaluation and speech continuation evaluation.

\subsubsection{Few-shot In-context Learning}

To systematically assess the overall capabilities of \mimoaudiobase after large-scale pretraining, we follow the GPT-3–style evaluation paradigm \citep{brown2020languagemodelsfewshotlearners} and adopt a \textbf{few-shot in-context learning} protocol for speech–text competence along three dimensions: modality-invariant general knowledge, auditory comprehension and reasoning, and speech-to-speech generation. Table \ref{tab:icl_settings} provides an overview of our few-shot in-context learning evaluation setup.

\paragraph{Modality-Invariant General Knowledge}We define modality-invariant general knowledge as the ability to access and express the same underlying knowledge regardless of input or output modality. To assess this across speech and text, we construct SpeechMMLU\footnote{\url{https://huggingface.co/datasets/XiaomiMiMo/SpeechMMLU}}
 by synthesizing the questions and options from the MMLU dataset~\citep{hendrycks2021measuringmassivemultitasklanguage} into speech while preserving their semantic content. The dataset is filtered by subject and length, resulting in a total of 8,549 entries across 34 subjects. We use a commercial TTS system with diverse voices for the synthesis. It consists of four parallel splits, enabling same-question cross-modal controls for evaluating knowledge across text-to-text, speech-to-text, text-to-speech, and speech-to-speech scenarios.

\begin{itemize}
    \item \textbf{Text-to-Text (T2T)}: Serves as a metric for retention of text capability and shows whether competence gained from text pretraining are diluted by speech–text pretraining; it also provides an upper-bound reference for speech performance.

    \item \textbf{Speech-to-Text (S2T)}: Compared with T2T, S2T quantifies the cross-modal cost of mapping a spoken question to its semantic form while producing a text answer on general-knowledge items.

    \item \textbf{Text-to-Speech (T2S)}: Relative to T2T, T2S probes the consistency and controllability of converting semantic content to spoken output on general-knowledge items.

    \item \textbf{Speech-to-Speech (S2S)}: S2S provides a comprehensive measure of the model’s integrated potential for end-to-end speech interaction on general-knowledge by completing the listen–think–speak loop.
\end{itemize}

\begin{table}[t]
\centering
\begin{tabular}{ccccc}
\toprule
\textbf{Capabilities} & \textbf{Dataset} & \textbf{Input Modality} & \textbf{Output Modality} & \textbf{\#Examples} \\
\midrule
\multirow{4}{*}{General Knowledge} & \multirow{4}{*}{SpeechMMLU} & Text & Text & 5 \\
 & & Speech & Text & 5 \\
 & & Text & Speech & 5 \\
 & & Speech & Speech & 5 \\
\midrule
Audio Understanding & MMAU & Audio+Text & Text & 5 \\
\midrule
Speech-to-Speech & Refer to Table~\ref{tab:s2s_tasks} & Speech & Speech & 16 \\
\bottomrule
\end{tabular}
\caption{Settings of few-shot in-context learning evaluation.}
\label{tab:icl_settings} 
\end{table}

\paragraph{Auditory Comprehension and Reasoning}While the S2T split of SpeechMMLU evaluates the model’s ability to recover semantics from speech and answer general-knowledge questions, it offers limited coverage of {non-semantic auditory factors}. To fully characterize MiMo-Audio’s upper bound in auditory understanding after large-scale speech–text pretraining, we extend the evaluation beyond basic semantic understanding to additional dimensions of the acoustic world. Accordingly, we assess the model on the MMAU test suite \citep{sakshi2024mmaumassivemultitaskaudio} under a few-shot in-context learning setup. MMAU comprises audio information extraction and reasoning QA across three domains: speech, environmental sounds, and music.

\paragraph{Speech-to-Speech Generation}\mimoaudioname represents speech with high-fidelity audio tokens that serve as a unified interface for perception and generation, thereby casting pretraining as high-fidelity compression over large-scale speech corpora. We hypothesize that sufficiently effective compression induces in-context learning ability that naturally generalizes to various downstream speech-to-speech tasks without parameter updates. To test this, we design a few-shot in-context speech-to-speech evaluation protocol that conditions exclusively on paired speech exemplars provided in context. Detailed descriptions of each speech-to-speech generation task can be found in Table \ref{tab:s2s_tasks}.

\begin{table}[ht]
\centering
\small
\setlength{\tabcolsep}{6pt}
\begin{tabularx}{\textwidth}{lXXX}
\toprule
\textbf{Task} & \textbf{Examples} & \textbf{Input} & \textbf{Expected Output} \\
\midrule
\textbf{Voice Conversion} &
Paired utterances from speakers A and B that share identical semantic content. &
Utterance from speaker A whose semantics differ from the examples. &
Utterance that preserves the input semantics but is rendered with speaker B's timbre. \\
\addlinespace[2pt]
\textbf{Emotion Conversion} &
Paired utterances from a fixed speaker with emotion A and emotion B; each pair shares identical semantics. &
Utterance from the same speaker with emotion A whose semantics differ from the examples. &
Utterance with the same timbre and semantics as the input but with emotion B. \\
\addlinespace[2pt]
\textbf{Rate Conversion} &
Paired utterances from a fixed speaker with rate A and rate B; each pair shares identical semantics. &
Utterance from the same speaker with rate A whose semantics differ from the examples. &
Utterance with the same timbre and semantics as the input but with rate B. \\
\addlinespace[2pt]
\textbf{Speech Denoising} &
Paired utterances from a fixed speaker including a noisy recording and its related clean version. &
Noisy utterance from the same speaker whose semantics differ from the examples. &
Denoised version of the input utterance. \\
\addlinespace[2pt]
\textbf{Speech Translation} &
Paired En-Zh utterances, with speakers not fixed across examples. &
English sentence to be translated. &
Translated Chinese sentence. \\
\bottomrule
\end{tabularx}
\caption{Example tasks for few-shot in-context speech-to-speech evaluation.}
\label{tab:s2s_tasks}
\end{table}



\subsubsection{Speech Continuation}

Continuation represents a fundamental capability of autoregressive language models. Through generative pretraining on extensive text corpora, text language models like GPT-3~\citep{brown2020languagemodelsfewshotlearners} acquire the ability to produce coherent textual continuations from input prompts. Analogously, \mimoaudioname undergoes generative pretraining on large-scale speech corpora and performs language modeling over high-fidelity audio tokens. This training paradigm endows the model with general speech continuation capabilities: given a brief speech prompt, \mimoaudiobase can generate semantically coherent continuations while preserving critical acoustic characteristics of the input, including: (i) speaker-specific characteristics such as identity and timbre, (ii) prosodic features encompassing rhythm, intonation, and tempo, (iii) environmental acoustics and non-speech audio elements (e.g., applause, laughter, sighs).

To probe this capability, we collect speech prompts from diverse domains, including stand-up comedy, public oratory, broadcast journalism, poetry recitation, audiobook narration, and academic lectures, as well as multi-speaker scenarios such as debates, interviews, and theatrical performances. 

\subsection{Results}

\begin{table}[h]
\centering
\resizebox{\linewidth}{!}{
\begin{tabular}{cc|cccc}
\toprule
\multicolumn{2}{c}{\multirow{2}{*}{\textbf{Task}}} & \textbf{Baichuan-Audio} & \textbf{Kimi-Audio} & \textbf{Step-Audio2-mini} & \textbf{\mimoaudioname} \\
& & 7B-Base & 7B-Base & 7B-Base & 7B-Base \\
\midrule
\multirow{4}{*}{SpeechMMLU} & S2S & 31.9 & 11.8 & 51.8 & \textbf{69.1} \\
 & S2T & 29.9 & 67.9 & 67.8 & \textbf{69.5} \\
 & T2S & 16.7 & 0.0 & 63.4 & \textbf{71.5} \\
 & T2T & 71.1 & 70.7 & \textbf{74.1} & 72.5 \\
\midrule
\multirow{4}{*}{MMAU} & Overall & 25.9 & 28.6 & 60.3 & \textbf{66.0} \\
 & Speech & 14.4 & 29.4 & 55.0 & \textbf{67.6} \\
 & Sound & 30.3 & 31.5 & \textbf{67.9} & 65.2 \\
 & Music & 32.9 & 24.8 & 58.1 & \textbf{65.3} \\
\bottomrule
\end{tabular}
}
\caption{Results on SpeechMMLU and MMAU. We compare \mimoaudiobase against Baichuan-Audio-Base~\citep{li2025baichuan}, Kimi-Audio-Base~\citep{kimi_audio}, and Step-Audio2-mini-Base~\citep{step_audio2}.}
\label{tab:main_results}
\end{table}

\paragraph{Emergent Ability}
As shown in Figure~\ref{fig:comparison}, we observed significant emergent abilities across multiple evaluation benchmarks, including 5-shot SpeechMMLU (T2S and S2S), 16-shot Voice Conversion, and 16-shot Speech-to-Speech Translation. During the initial training stage (before the data volume reached approximately 0.7 trillion tokens), the model's performance on these tasks was negligible, indicating it had not yet acquired the atomic skills required to solve these complex problems. However, once the training volume surpassed this critical threshold, the model's performance underwent a sharp, non-linear surge, exhibiting a characteristic "phase transition." Following this leap, performance continued to improve steadily before eventually stabilizing, indicating that the model had fully mastered and consolidated this new ability.

This emergence of capabilities from a near-zero baseline, rather than through gradual improvement, is a direct manifestation of the model autonomously developing advanced generalization abilities through large-scale learning. This finding strongly supports our assertion that this represents a "GPT-3 moment" for the speech domain: through sufficiently large-scale, lossless compression-based pre-training, models can spontaneously learn to solve complex, previously unseen tasks, thereby achieving task generalization.

\paragraph{Speech Intelligence}
MiMo-Audio model delivered exceptional performance in speech intelligence tasks, with its superiority primarily manifested in two key dimensions: its SpeechMMLU score and the magnitude of its "modality gap".

We use SpeechMMLU score to measure a model’s capacity to perform complex reasoning and knowledge-based question-answering directly with speech as input or output. As shown in Table~\ref{tab:main_results}, MiMo-Audio achieves the highest scores in both SpeechMMLU-S2S~(69.1), SpeechMMLU-S2T~(69.5) and SpeechMMLU-T2S~(71.5). Step-Audio2 mini-base achieved a relatively competitive score in S2T~(67.8) but its performance decrease to 51.8 in S2S, revealing significant fluctuations across different speech tasks. Kimi-Audio-base fared moderately in S2T~(67.9) yet exhibited a critical weakness in S2S. Baichuan-Audio-base, meanwhile, posted consistently low scores in both tasks~(31.9 and 29.9). MiMo-Audio thus emerged as the only evaluated model capable of sustaining high-level performance across all speech reasoning tasks.

The modality gap, a metric gauging the consistency of a model’s capabilities between speech and text modalities, is calculated as the difference between a model’s text2text score and its speech2speech (S2S) score. MiMo-Audio’s modality gap is 3.4 points, while Step-Audio2 mini-base’s gap stands at 22.3 points, Kimi-Audio-base’s at 58.9 points, and Baichuan-Audio-base’s at 39.2 points. The data confirms that MiMo-Audio boasts the smallest modality gap among all models, which underscores that its architectural design is uniquely effective at preserving the continuity of core reasoning capabilities across distinct input modalities.

\paragraph{General Audio Understanding}
As shown in Table~\ref{tab:main_results}, MiMo-Audio demonstrated the superior general audio understanding capabilities among current open-source models. This advantage is reflected not only in its overall score but also in its balanced performance across all subtasks.

In terms of the MMAU overall score, MiMo-Audio achieved 66.0 points, which is 5.7 points higher than Step-Audio2 mini-base~(60.3 points), the second-place model. Compared to Kimi-Audio-base (28.6 points) and Baichuan-Audio-base (25.9 points), MiMo-Audio’s score is significantly higher. This lead in the total score intuitively reflects its overall performance superiority.

General audio understanding requires models to perform well across diverse audio types, and MiMo-Audio excels with a balanced capability distribution. It achieved consistently high scores across three subdomains: speech (67.6), sound effects (65.2), and music (65.3), with no obvious performance shortcomings. In contrast, while Step-Audio2 mini-base obtained the highest score in sound effects~(67.9), it performed relatively poorly in speech~(55.0) and music~(58.1). The Kimi-Audio-base and Baichuan-Audio-base models, meanwhile, scored consistently lower across all subtasks.

\paragraph{Speech Task Generalization}
Figure~\ref{fig:comparison} reports results for voice conversion and speech-to-speech translation under the 16-shot in-context learning setting. For other speech-to-speech generation tasks that are less amenable to automatic evaluation, we present qualitative demos\footnote{\url{https://xiaomimimo.github.io/MiMo-Audio-Demo}}. We strongly encourage readers to visit the demo page and listen to results. In Figure~\ref{fig:comparison}, few-shot prompting reveals that the abilities of general speech-to-speech generation and modality-invariant general knowledge (SpeechMMLU, T2S/S2S) emerge together at similar training scales. This alignment suggests a shared underlying speech competence is emerging, enabling MiMo-Audio to generalize to controlled transformations of fine grained factors such as speaker identity, emotion, and speaking rate.


\paragraph{Speech Continuation}
We strongly recommend visiting the demo page to listen to our speech continuation demos. As showcased on our demo page, across these varied contexts, MiMo-Audio-Base can perform speech continuation for different scenarios—including Game Live Streaming, Teaching, Recitation, Singing, Talk Show, and Debate—generating speech that features coherent semantics, natural prosodic connection, consistent acoustic conditions, and scene relevance, without requiring any parameter adaptation. Specifically, for singing speech, it can generate consistent and pleasant vocal melodies; for talk show continuation, it can even produce audience cheers at appropriate moments; for two-person debate continuation, it can generate two-person speech with consistent viewpoints, coherent semantics, and smooth prosody; for dialect speech continuation, it can generate content with consistent accents; for scenarios such as game live streaming and teaching, it can generate highly expressive and colloquial speech, with volume variations and colloquial expressions like stutters added at appropriate times; and for recitation speech continuation, it can generate emotional speech with professional recitation quality.
These results indicate that through generative pretraining on large-scale, naturalistic audio recordings, MiMo-Audio-Base has acquired comprehensive and generalizable audio knowledge, demonstrating its potential for broader audio understanding and generation applications.

\section{Post-Training}

\subsection{Data}

\label{sec:410_postrain_data}

The objective of our post-training data strategy is to use a series of supervised instruction fine-tuning datasets to activate the pre-trained model’s understanding and generation capabilities on different tasks.

\subsubsection{Audio Understanding}
To activate the model's audio understanding and reasoning capabilities, we integrated multiple open-source datasets covering speech, sounds, and music. To address the problems of label noise and singular task paradigms within the data, we designed a LLM-based pipeline for data cleaning and augmentation. This ultimately generated a large amount of diverse audio understanding data, such as audio captioning and audio question answering.

\subsubsection{Speech Generation}
To activate the model's speech generation capabilities, we extracted a high-quality speech subset from the pre-training data and constructed instruction data based on audio captions. The model is required to generate matching audio according to this instruction. This training method is intended to strengthen the model's instruction-following capability and achieve controllable, high-quality speech generation.

\subsubsection{Spoken Dialogue}
To activate the  model's ability to generate speech with diverse styles and high expressiveness in different dialogue scenarios, we constructed a massive spoken dialogue dataset containing single-turn and multi-turn conversations.
These spoken dialogues consist of user queries and assistant replies. The content is primarily sourced from rigorously screened text data to ensure reliable quality.

To make \mimoaudioname adapt to diverse conversational styles, we first perform stylistic rewriting on the colloquially adapted question-answer pairs. We then use the in-house MiMo-TTS system to synthesize speech with appropriate style and emotion. During synthesis, we randomly select prompt audio from a voice library containing a large number of timbres to ensure coverage of different vocal expressiveness.

\begin{table}[t]
\centering
\small
\begin{tabularx}{0.9\textwidth}{>{\centering\arraybackslash}X c c c}
\toprule
\textbf{Task Type} & \textbf{Dataset} & \textbf{Input Modality} & \textbf{Output Modality}\\
\midrule
\multirow{2}{*}{\parbox{2.5cm}{\centering ASR}} & AISHELL1 & Speech & Text \\
                     & LibriSpeech test-clean & Speech & Text \\
\midrule

\multirow{4}{*}{\parbox{3.5cm}{\centering TTS}} & SeedTTS test-Zh & Text & Speech \\
                     & SeedTTS test-En & Text & Speech \\
                     & InstructTTSEval-Zh & Text & Speech \\
                     & InstructTTSEval-En & Text & Speech \\
\midrule

\multirow{4}{*}{\parbox{3.5cm}{\centering Audio Understanding \\ and Reasoning}} & MMSU & Speech+Text & Text \\
                     & MMAU & Audio+Text & Text \\
                     & MMAR & Audio+Text & Text \\
                     & MMAU-Pro & Audio+Text & Text \\
\midrule

\multirow{4}{*}{\parbox{3.5cm}{\centering Spoken Dialogue}} & Big Bench Audio S2T & Speech & Text \\
                     & Big Bench Audio S2S & Speech & Speech \\
                     & MultiChallenge Audio S2T & Speech & Text \\
                     & MultiChallenge Audio S2S & Speech & Speech \\
\bottomrule
\end{tabularx}
\caption{Evaluation Settings of \mimoaudioinst.}
\label{tab:sft_settings}
\end{table}

\subsection{Training}
In the post-training stage, all model parameters, including the patch encoder, LLM, and patch decoder, are fine-tuned. For this, we curated a comprehensive training dataset of 100 billion tokens, encompassing six distinct task formats: ASR, TTS, audio understanding, spoken dialogue, instruction-following TTS, and text dialogue. While data for ASR, TTS, and text dialogue are sourced from open-source collections, the remaining tasks utilized the high-quality datasets detailed in Section~\ref{sec:410_postrain_data}.

For speech generation and spoken dialogue tasks, we continue to employ the text-guided interleaving strategy from the second pre-training stage, where the model interleaves text tokens and speech patches in a fixed 5:5 ratio. The loss weights are also kept consistent with this stage: 100 for text tokens and 12, 8, 6, 4, 2, 2, 1, 1 for audio tokens. As specified in Table~\ref{table:training_config}, we set the learning rates for the patch encoder and decoder to 5e-5 and the LLM to 1e-5, respectively, with a cosine decay schedule. The model is trained with a context length of 8192 and a batch size of 2.1M tokens.

\begin{table}[htp]
\centering
\resizebox{\linewidth}{!}{
\begin{tabular}{ccc}
\toprule
\textbf{Datasets} & \textbf{Model} & \textbf{Performance} \\
\midrule
\multicolumn{3}{c}{\textit{Audio Understanding}}\\
\midrule
\multirow{8}{*}{\makecell{\textbf{MMAU}\\Speech | Sound | Music | Overall}}&\mimoaudioinst&68.47 | \underline{\textbf{82.58}} | \underline{\textbf{73.65}} | \underline{\textbf{74.90}}\\
 &\geminiflash &\textbf{76.58} | 73.27 | 65.57 | 71.80\\
 &\afthree & 66.37 | 79.58 | 66.77 | 73.30\\
 & \stepaudioinst & 68.16 | 79.30 | 68.44 | 72.73\\
 & \kimiaudioinst & 62.16 | 75.68 | 66.77 | 68.20\\
 & \qwenomni& \underline{\textbf{70.60}} | 78.10 | 65.90 | 71.50\\
 &\glmvoice & 35.44 | 27.63 | 27.84 | 30.30\\
 \midrule
\multirow{9}{*}{\textbf{MMAU-Pro}}&\mimoaudioinst&\underline{\textbf{53.35}}\\
 &\geminiflash &\textbf{59.20}\\
 &\afthree &51.70\\
 & \stepaudioinst & 47.91\\
 & \kimiaudioinst & 46.60\\
 & \qwenomni& 52.20\\
 &\glmvoice & 38.25\\
 &GPT-4o-Audio & 52.50\\
 \midrule
\multirow{9}{*}{\textbf{MMAR}}&\mimoaudioinst&\underline{\textbf{63.60}}\\
 &\geminiflash &\textbf{65.60}\\
 &\afthree &58.50\\
 & \stepaudioinst & 55.80\\
 & \kimiaudioinst & 48.00\\
 & \qwenomni& 56.70\\
 &\glmvoice & 29.50\\
 &GPT-4o-Audio & 63.50\\
 \midrule
 \multirow{9}{*}{\makecell{\textbf{MMSU}\\Perception | Reasoning | Overall }}&\mimoaudioinst&\underline{\textbf{46.86}} | 76.98 | \underline{\textbf{61.70}}\\
 &\mimoaudioinst+Think&\underline{\textbf{51.71}} | 74.79 | \underline{\textbf{62.88}}\\
 &Gemini 1.5 Pro &\quad  - \quad | \quad -  \quad | 60.70\\
 &\afthree & \quad  - \quad | \quad  -  \quad | 61.40\\
 & \stepaudioinst &42.71 | 72.60 | 57.18\\
 & \kimiaudioinst &44.84 | 75.70 | 59.78\\
 & \qwenomni &42.67 | \underline{\textbf{77.64 }}| 58.10\\
 &\glmvoice &11.04 | 16.16 | 13.30\\
 \midrule
 \multicolumn{3}{c}{\textit{Spoken Dialogue}}\\
 \midrule
\multirow{6}{*}{\makecell{\textbf{Big Bench Audio }\\S2T | S2S }}&\mimoaudioinst& \underline{\textbf{72.90}} | \underline{\textbf{60.20}}\\
 &\gptaudio &70.20 | \textbf{67.20}\\
 & \stepaudioinst &50.90 | 47.50\\
 & \kimiaudioinst &59.40 | 51.00\\
 & \qwenomni &54.20 | 53.60\\
 &\glmvoice &44.80 | 42.70\\
  \midrule
\multirow{6}{*}{\makecell{\textbf{MultiChallenge Audio }\\S2T | S2S }}&\mimoaudioinst& \underline{\textbf{15.15}} | \underline{\textbf{10.10}}\\
 & \stepaudioinst &13.64 | 8.08\\
 & \kimiaudioinst &7.07 | 1.01\\
 & \qwenomni &11.11 | 8.08\\
 &\glmvoice &9.09 | 6.06\\
\bottomrule
\end{tabular}
}
\caption{Results on audio understanding and spoken dialogue benchmarks. \textbf{Bold} indicates the best performance overall, and \underline{\textbf{underline}} marks the best among open-source models. +Think indicates turning on thinking.}
\label{tab:post_main_results}
\end{table}

\begin{table}[htp]
\centering
\resizebox{\linewidth}{!}{
\begin{tabular}{ccc}
\toprule
\textbf{Datasets} & \textbf{Model} & \textbf{Performance} \\
\midrule
\multicolumn{3}{c}{\textit{TTS}}\\
\midrule
\multirow{2}{*}{\makecell{\textbf{Seed-TTS-Eval }\\ZH | EN | ZH-Hard }}&\mimoaudioinst& \textbf{1.96} | 5.37 | \textbf{14.14} \\
 & \stepaudioinst &2.13 | \textbf{3.18} | 16.31\\
 \midrule
 \multicolumn{3}{c}{\textit{Instruct-TTS}}\\
\midrule
\multirow{2}{*}{\makecell{\textbf{InstructTTSEval-EN}\\APS | DSD | RP | Overall}}&\mimoaudioinst& \textbf{80.60} | \textbf{77.63} | \textbf{59.54} | \textbf{72.59} \\
 & GPT-4o-mini-tts & 76.40 | 74.30 | 54.80 | 68.50\\
 \midrule
 \multirow{2}{*}{\makecell{\textbf{InstructTTSEval-ZH}\\APS | DSD | RP | Overall}}&\mimoaudioinst& \textbf{75.74} | \textbf{74.3} | \textbf{61.54} | \textbf{70.52} \\
 & GPT-4o-mini-tts & 54.90 | 52.30 | 46.0 | 51.07\\
 \midrule
 \multicolumn{3}{c}{\textit{ASR}}\\
 \midrule
 \multirow{3}{*}{\makecell{\textbf{ASR }\\Librispeech-test-clean | AISHELL }}&\mimoaudioinst& 3.50 | 1.65\\
 & \stepaudioinst &\textbf{1.87} | 0.95\\
 & \kimiaudioinst & 2.13 | \textbf{0.62} \\
\bottomrule
\end{tabular}
}
\caption{Results on the ASR and TTS benchmarks.}
\label{tab:tts_asr}
\end{table}

\subsection{Evaluation}

After post-training, we conducted a systematic evaluation of \mimoaudioinst, covering audio understanding, spoken dialogue, as well as speech recognition and generation. The specific configurations for each task type are shown in Table~\ref{tab:sft_settings}. In the following sections, we provide a detailed description of each task.

\subsubsection{Audio Understanding}
As a general-purpose audio model, we first assess the model's general audio understanding capabilities. Firstly, we adopt the MMSU~\citep{wang2025mmsumassivemultitaskspoken} benchmark, which focuses on multi-task spoken understanding. In addition to speech, we extend the evaluation to broader audio understanding tasks involving sound and music, using the MMAU~\citep{sakshi2025mmau} benchmark. To further assess the model’s audio reasoning capabilities, we also use MMAR~\citep{ma2025mmarchallengingbenchmarkdeep} and MMAU-Pro~\citep{kumar2025mmauprochallengingcomprehensivebenchmark}, which evaluate the model’s capacity to handle mixed audio inputs, such as speech, music, and environmental sounds, as well as its grasp of audio knowledge.

\subsubsection{Spoken Dialogue}

Speech interaction is one of the most crucial modalities for human–computer communication. To evaluate how well an audio-language model can follow user instructions and complete tasks in multi-turn dialogues, following OpenAI\footnote{\url{https://openai.com/index/introducing-gpt-realtime/}}, we first assess the model’s performance on Big Bench Audio\footnote{\url{https://huggingface.co/datasets/ArtificialAnalysis/big_bench_audio}}~\citep{srivastava2022beyond, suzgun2022challenging}
, a benchmark designed to measure the intelligence level of audio-language models. The response quality scores are derived from GPT-based evaluations. For spoken responses, the audio is first transcribed using the Whisper-Large-V3 \citep{10.5555/3618408.3619590} model and then evaluated by GPT-4o-mini.

Next, to evaluate how well the model can handle more complex dialogues, we use the MultiChallenge \citep{sirdeshmukh2025multichallengerealisticmultiturnconversation} dataset. This dataset requires models to generate appropriate responses for the final turn, based on the preceding dialogue history.

Since MultiChallenge was originally a text-based multi-turn interaction benchmark, we convert it into a speech-based version through the following steps:

\begin{itemize}
\item Filter out samples containing excessive mathematical symbols, tables, URLs, or other non-spoken formats.
\item Convert the remaining samples into speech using a commercial TTS model. Utterances from the same speaker within a sample are synthesized with a consistent voice, selected from a pool of 250 voices.
\end{itemize}

This results in two speech versions of MultiChallenge Audio: S2T (speech-to-text) and S2S (speech-to-speech). In the S2T version, the dialogue history is presented as text, while in S2S, it is entirely in speech.

\subsubsection{Speech Recognition and Generation}

As a native audio-language model, speech recognition and speech generation form the foundation for enabling more advanced speech tasks. To this end, we compare \mimoaudioinst with other audio-language models \citep{qwen_omni, kimi_audio, step_audio2} on automatic speech recognition (ASR) and text-to-speech (TTS) tasks.

For ASR, we evaluate using the widely adopted LibriSpeech \citep{7178964} test-clean set for English and the AISHELL-1 \citep{8384449} test set for Chinese. The ASR task is evaluated using Word Error Rate (WER) as the metric.

Beyond recognition capabilities, we also evaluate the speech generation ability of~\mimoaudioinst. We first assess the TTS performance of \mimoaudioinst on the SeedTTS~\citep{anastassiou2024seed} benchmark, which includes both English and Chinese subsets, as well as a more challenging hardcase subset for Chinese. In addition to conventional TTS evaluations, we conduct more advanced assessments on the InstructTTSEval \citep{huang2025instructttsevalbenchmarkingcomplexnaturallanguage} benchmark, which measures the ability of models to follow complex natural-language style control instructions to syntheses the corresponding speech, thereby jointly evaluating fidelity and expressive generation. For TTS tasks, we adopt WER as a basic evaluation metric, where synthesized speech is first transcribed by an ASR model \citep{10.5555/3618408.3619590, gao2023paraformerfastaccurateparallel} and then compared against the reference text. Moreover, InstructTTSEval leverages Gemini-based scoring to further assess the alignment between the generated speech and the input instructions.







\subsection{Results}

\paragraph{Audio Understanding}

For the audio understanding tasks, as shown in Table~\ref{tab:post_main_results}, the results on the MMSU and MMAU benchmarks demonstrate that \mimoaudioinst achieves leading performance in speech, audio, and music question answering. The overall scores on these two benchmarks outperform all open-source models, as well as closed-source models like Gemini 2.5 Flash and Gemini 1.5 Pro. 

For more challenging audio reasoning tasks, \mimoaudioinst also leads on the MMAU-Pro and MMAR benchmarks, achieving results that are close to Gemini 2.5 Flash. These results collectively demonstrate that \mimoaudioinst is a general-purpose and powerful audio understanding model.

\paragraph{Spoken Dialogue}
As shown in the Table~\ref{tab:post_main_results}, MiMo-Audio-7B-Instruct achieves the best performance among all open-source models across both the Big-Bench-audio and Multi-Challenge-Audio tasks, and its results are close to those of the proprietary model gpt-4o. On the Big-Bench-audio benchmark, MiMo-Audio-7B-Instruct scores 72.90 (S2T) and 60.20 (S2S), ranking second only to gpt-4o while significantly outperforming all other open-source models. Similarly, on the Multi-Challenge-Audio benchmark, it achieves 15.15 (S2T) and 10.10 (S2S), again leading the open-source group by a notable margin.
In summary, MiMo-Audio-7B-Instruct not only outperforms all other open-source models by a wide margin, but also narrows the gap with the state-of-the-art proprietary model gpt-4o, demonstrating strong competitiveness and practical potential.
We encourage you to visit our demo page\footnote{\url{https://xiaomimimo.github.io/MiMo-Audio-Demo}} to explore our speech-to-speech dialogue demos. Our model demonstrates strong human-likeness and expressive conversational abilities, along with solid performance in knowledge understanding, emotional intelligence, dialogue skills, and instruction following. It also supports dialects and multilingual communication.

\paragraph{Speech Recognition and Generation}
As shown in Table~\ref{tab:tts_asr}, MiMo-Audio-7B-Instruct demonstrates strong performance in both ASR and TTS tasks among open-source large speech models. On the ASR and TTS benchmarks, it achieves similar results to other open-source models such as Step-Audio2-mini and Kimi-Audio-Instruct. In the InstructTTS evaluation, MiMo-Audio-7B-Instruct outperforms gpt-4o-mini-tts on both English and Chinese subsets, with especially competitive results on overall metrics.
These results highlight MiMo-Audio-7B-Instruct's effectiveness in controllable text-to-speech generation, positioning it as a leading open-source solution in this space.

\section{Conclusion}

In this work, we have demonstrated that scaling next-token prediction pre-training on massive-scale, lossless audio data is a viable path toward achieving general-purpose speech intelligence. By pre-training on an unprecedented corpus of over 100 million hours, \mimoaudioname successfully transcends the limitations of task-specific fine-tuning that characterize existing audio language models.

Our primary contribution is the empirical validation that a "GPT-3 moment" is achievable in the speech domain. We observed the distinct emergence of powerful few-shot learning capabilities after crossing a critical data threshold, enabling the model to generalize to a wide array of tasks—including complex voice conversion, style transfer, and speech editing—without task-specific training. Furthermore, we presented a comprehensive blueprint for this paradigm, encompassing a novel unified high-fidelity audio tokenizer, a scalable architecture, and a phased training strategy. \mimoaudioinst achieves state-of-the-art performance on multiple benchmarks and rivals closed-source systems.

Ultimately, this research provides a foundational methodology for building truly versatile audio language models. We believe this work marks a significant step towards creating more natural, flexible, and intelligent systems that can understand and generate speech with human-like adaptability.

\section{Limitations and Future Work}
\paragraph{Limited In-Context-Learning Performance}
The in-context learning capability of MiMo-Audio-Base remains constrained. While the pre-trained model can fulfill a variety of novel tasks beyond the scope of its pre-training via in-context learning, it exhibits suboptimal performance in certain scenarios—such as speech generation with background music and the processing of complex sound events. Moving forward, we aim to enhance MiMo-Audio’s capability in general audio generation.

\paragraph{Unstable Spoken Dialogue Performance}
MiMo-Audio-Instruct demonstrates several limitations in speech dialogue, including timbre discontinuities, unstable audio quality, mispronunciations, and inconsistent compliance with system prompts. Notably, it is highly prone to mispronouncing complex symbols and formulas, and its style control during dialogue is also unstable. In future work, we will leverage reinforcement learning (RL) to improve the stability of the model’s performance.

\paragraph{Limited Thinking Performance}
When integrating the thinking mechanism, MiMo-Audio-Instruct yields performance improvements exclusively in speech-related understanding tasks, whereas it induces performance degradation in sound and music understanding tasks. Our analysis of failure cases (bad cases) reveals that this phenomenon stems from hallucinations introduced by the model during the thinking process. Going forward, we plan to enhance the model’s audio understanding capability through reinforcement learning (RL).

\bibliography{main}


\appendix
\newpage

\section{Contributions and Acknowledgments}
We would like to express our sincere gratitude to all contributors for their invaluable support and efforts, including the Xiaomi LLM-Plus, NGK, MiChat, Mify, Data Platform and CloudML teams, as well as those not explicitly listed in this paper.
\textit{Authors within each role are listed alphabetically by their first name}.

\definecolor{ourblue}{RGB}{0, 0, 0}
\definecolor{ourgreen}{RGB}{0, 0, 0}
\definecolor{ourred}{RGB}{0, 0, 0}

\begin{multicols}{2} %
\noindent
\textbf{\color{ourred} Core Contributors} \\
\color{ourred} Dong Zhang \\
\color{ourred} Gang Wang \\
\color{ourred} Jinlong Xue   \\
\color{ourred} Kai Fang \\
\color{ourred} Liang Zhao \\
\color{ourred} Rui Ma \\
\color{ourred} Shuhuai Ren \\
\color{ourred} Shuo Liu \\
\color{ourred} Tao Guo   \\
\color{ourred} Weiji Zhuang \\
\color{ourred} Xin Zhang \\
\color{ourred} Xingchen Song \\
\color{ourred} Yihan Yan \\
\color{ourred} Yongzhe He   \\
\color{ourred} Cici\textsuperscript{\dag} \\

\noindent
\textbf{\color{ourblue} Deployment \& Evaluation} \\
\color{ourblue} Bowen Shen \\
\color{ourblue} Chengxuan Zhu   \\
\color{ourblue} Chong Ma \\
\color{ourblue} Chun Chen   \\
\color{ourblue} Heyu Chen   \\
\color{ourblue} Jiawei Li \\
\color{ourblue} Lei Li \\
\color{ourblue} Menghang Zhu \\
\color{ourblue} Peidian Li \\
\color{ourblue} Qiying Wang   \\
\color{ourblue} Sirui Deng \\
\color{ourblue} Weimin Xiong   \\
\color{ourblue} Wenshan Huang   \\
\color{ourblue} Wenyu Yang \\
\color{ourblue} Yilin Jiang   \\
\color{ourblue} Yixin Yang   \\
\color{ourblue} Yuanyuan Tian \\
\color{ourblue} Yue Ma   \\
\color{ourblue} Yue Yu \\
\color{ourblue} Zihan Zhang \\
\color{ourblue} Zihao Yue \\

\vspace{0.6em}
\hrule
\vspace{0.4em}

{\footnotesize
\textsuperscript{\dag} Corresponding author
}

\noindent
\textbf{\color{ourblue} Additional Contributors} \\
\color{ourred} Bangjun Xiao   \\
\color{ourred} Bingquan Xia \\
\color{ourred} Bofei Gao \\
\color{ourred} Bowen Ye \\
\color{ourred} Can Cai \\
\color{ourred} Chang Liu \\
\color{ourred} Chenhong He \\
\color{ourred} Chunan Li   \\
\color{ourred} Dawei Zhu \\
\color{ourred} Duo Zhang \\
\color{ourred} Fengyuan Shi   \\
\color{ourred} Guoan Wang \\
\color{ourred} Hailin Zhang \\
\color{ourred} Hanglong Lv \\
\color{ourred} Hanyu Li   \\
\color{ourred} Hao Tian \\
\color{ourred} Heng Qu \\
\color{ourred} Hongshen Xu \\
\color{ourred} Houbin Zhang   \\
\color{ourred} Huaqiu Liu \\
\color{ourred} Jiangshan Duo   \\
\color{ourred} Jianguang Zuo \\
\color{ourred} Jianyu Wei   \\
\color{ourred} Jiebao Xiao \\
\color{ourred} Jinhao Dong \\
\color{ourred} Jun Shi \\
\color{ourred} Junhao Hu   \\
\color{ourred} Kainan Bao \\
\color{ourred} Kang Zhou  \\
\color{ourred} Linghao Zhang   \\
\color{ourred} Meng Chen \\
\color{ourred} Nuo Chen \\
\color{ourred} Peng Zhang \\
\color{ourred} Qianli Chen   \\
\color{ourred} Qiantong Wang \\
\color{ourred} Rang Li   \\
\color{ourred} Shaohui Liu \\
\color{ourred} Shengfan Wang   \\
\color{ourred} Shicheng Li \\
\color{ourred} Shihua Yu \\
\color{ourred} Shijie Cao   \\
\color{ourred} Shimao Chen \\
\color{ourred} Shuhao Gu \\
\color{ourred} Weikun Wang \\
\color{ourred} Wenhan Ma \\
\color{ourred} Xiangwei Deng \\
\color{ourred} Xing Yong \\
\color{ourred} Xing Zhang \\
\color{ourred} Xu Wang \\
\color{ourred} Yifan Song \\
\color{ourred} Yihao Zhao   \\
\color{ourred} Yingbo Zhao \\
\color{ourred} Yizhao Gao   \\
\color{ourred} Yu Cheng   \\
\color{ourred} Yu Tu \\
\color{ourred} Yudong Wang \\
\color{ourred} Zhaojun Huang   \\
\color{ourred} Zhengju Tang   \\
\color{ourred} Zhenru Lin \\
\color{ourred} Zhichao Song \\
\color{ourred} Zhipeng Xu   \\
\color{ourred} Zhixian Zheng \\
\color{ourred} Zihan Jiang \\

\end{multicols} %


\end{document}